# QUEST: Quadriletral Senary bit Pattern for Facial Expression Recognition

Monu Verma[1]
monuverma.cv@gmail.com

Prafulla Saxena[2]
prafulla1308@gmail.com

S. K. Vipparthi[3]
skvipparthi@mnit.ac.in

Gridhari Singh[4]
gsingh.cse@mnit.ac.in

Dept. of Computer Science and Engineering, Malaviya national Institute of Technology, Jaipur, India

*Abstract*— Facial expression has significant role to analyzing human cognitive state. Deriving an accurate facial appearance representation is critical task for an automatic facial expression recognition application. This paper provides a new feature descriptor named as Quadrilateral Senary bit Pattern for facial expression recognition. The QUEST pattern encoded the intensity changes by emphasizing relationship between neighboring and reference pixels by dividing them into two quadrilaterals in local neighborhood. Thus, the resultant gradient edges reveal the transitional variation information, that improves the classification rate by discriminating expression classes. Moreover, it also enhances the capability of the descriptor to deal with view point variations and illumination changes. The trine relationship in quadrilateral structure helps to extract the expressive edges and suppressing noise elements to enhance the robustness to noisy conditions. The QUEST pattern generates a six-bit compact code, which improve the efficiency of the FER system with more discriminability. The effectiveness of proposed method is evaluated by conducting several experiments on four benchmark datasets: MMI, GEMEP-FERA, OULU-CASIA and ISED. The experimental results show better performance of the proposed method as compared to existing state-art-the approaches.

## I. INTRODUCTION

The emotional state of a human can be inferred through the various modalities such as gestures, biophysiological data and facial expressions. Facial expressions convey relevant informational cues about the mental state of a person. Consequently, the facial expressions deliver more accurate perception to the judgment of human profiling as compare to other agents such as body gestures, voice and biophysiological data. Therefore, the automatic facial analysis appears in different fields such as human-computer interaction, marketing, online tutoring, depression analysis, surveillance, entertainment and low enforcement etc. Since, automatic facial expression recognition (FER) has attracted peoples towards itself.

Ekman [1] presented a detailed study about the facial expression and discovered six basic expression: anger, disgust, fear, happy, sad and surprise. Furthermore, to analyze the facial expressions more efficiently, Ekman and Friesen [2] developed a facial action coding system (FACS) by dividing a face into smaller units, named as action units (AUs). These action units interpret the micro level changes in facial muscles. Thus, the combination of different AUs along with different intensities defines each expression. The main challenges towards design a robust automatic FER system are head pose variation, illumination changes, ethnicity, gender and age differences etc.

General framework of FER system follows three steps: face acquisition, salient feature extraction and expression classification. In face acquisition phase, facial region is extracted and cropped to eliminate the background noise. Furthermore, salient features are extracted and fed into the classifier for expression classification. The efficiency of a FER system is mainly depending on the feature extraction techniques. Feature extraction techniques describe the facial appearance changes. With using some descriptor, the muscle variation on face is captured and features are extracted. Inadequate feature extraction will degrade the performance of the FER system, regardless of the good classification technology. Therefore, it is essential to design a robust feature descriptor to extract facial appearance variations. There are two common categories to represent the expression features: predesigned and learned features. Generally, predesigned features are extracted by encompassing handcrafted filters and then, response features are passed to the classification phase to discriminate the expression among six classes. Predesigned feature extraction techniques also categorized into two categories: geometric shape based and facial appearance-based methods [3]. The former technique [4-5] represents the facial appearance by encoding the shape and location related to the facial points. However, the effectiveness of the FER system fully relies on the accurate fiducial points, which are hard to detect under variant conditions such as view point and ethnicity variations. The appearance-based techniques resolve this issue by extracting the salient features by applying filters on the facial images. These filters can be masked over the whole image (holistic) or specific image region (local) to represent the invariant features of the face. The appearance-based methods show excellent performance as they captured accurate set of features. Many approaches based on statistics and transformation techniques has been reported in [6-8]. Recently, local descriptors are widely accepted due to their robustness regarding change in lighting and pose. In literature there are many local descriptors has been proposed such as, Gabor features [9], Elastic Bunch Graph Matching [10] and Local Binary Pattern (LBP) [11]. LBP has gained popularity as it yields better accuracy rate as compare to previous approaches. Conversely, LBP faced problem in variant environments such as illumination changes and noise. To enhance the performance of LBP, many extensions of LBP has been introducing like Local Ternary Pattern (LTP) [12], Two Phase LBP [13], and Local Phase Quantization (LPQ) [14]. Furthermore, Jabid et al. [15] proposed local directional pattern (LDP), that encodes the directional information to represent the facial expression features. The directional information is calculated by utilizing eight compass masks. However, the performance of the directional pattern is excellent with the constrained factors but their performance degrades with variant conditions. Furthermore, Rivera et al. [16] designed a principle direction number-based descriptor, local directional number pattern (LDN) to improve the robustness regarding illumination and noisy environment. After that, they also proposed local directional texture pattern (LDTexP) [17], by combining advantages of texture and directional information. Ryu et al. [18]

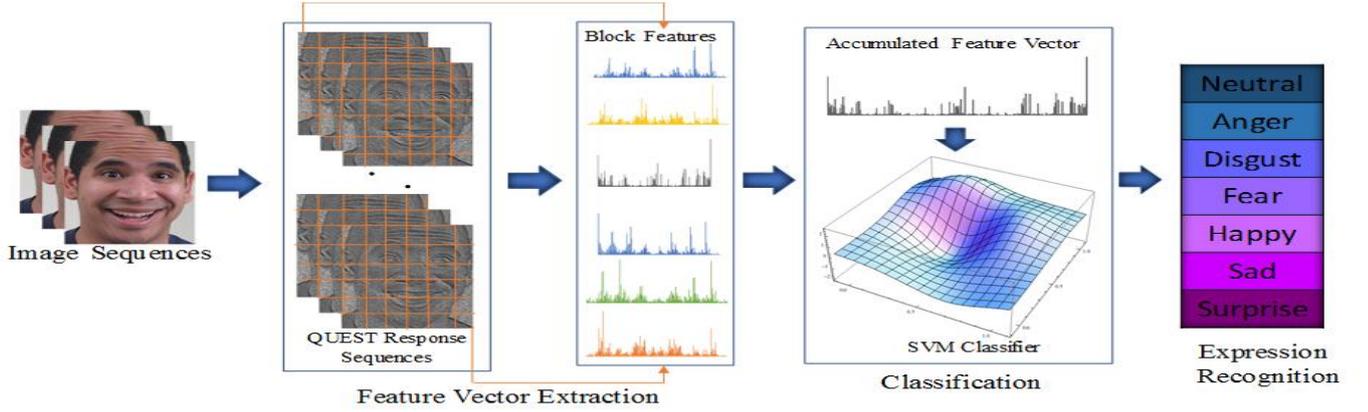

Fig. 1. Working model of the proposed facial expression recognition system. For image sequence, first the region of interest is detected and normalized. Then, we generate QUEST pattern and extract the response feature vectors. Finally, resultant features are accumulated and poured to classifier for the classification.

introduced local directional ternary pattern (LDTerP) to encode the salient features by extracting active patches from the facial image. Though, learned features state that imperative features are automatically discovered with the help of training images using some neural network architecture.

Now a day, convolution neural network (CNN) have also been used to extract the salient features from the facial expressions. There are many architectures has been proposed [19-21] to increase the recognition rate of the FER system. These architectures have the ability to automatically learn the relevant features from the training image set and classify the input image to the labeled expression class. Khorami et al. [22] designed a three-layer CNN architecture to learn salient features from the AUs regions. Furthermore, Lopes et al. [23] generated a big image set through preprocessing to improve the robustness of CNN network. Ding et al. [24] created a two-phase architecture: in the first phase, the network is trained for the expressive facial regions using the Face2Net model and in second phase, a fully connected layer network is added to classify the response feature vector.

In this paper, we proposed a new feature descriptor QUEST by extracting 3-bit relations in a quadrilateral structure at a given location. The QUEST characterizes a local region in two quads and encodes senary bit pattern. This pattern able to extricate in the facial appearance in a region and also enhances the robustness to view point and illumination variations. The working framework of the proposed FER system using QUEST descriptor is shown in Fig. 1

The rest of the paper is organized as follows: In Section II, detailed discussion of the proposed method is presented with visual representation. In Section III, the experimental results and comparative analysis are discussed. We conclude our paper in Section V.

## II. THE PROPOSED METHOD

In this paper, we proposed QUEST descriptor for facial expression recognition system. The QUEST encodes the senary bit pattern by embodying trine relations in the quadrilateral structure to represent the texture variations in the local region. These trine relations are generated from each quadrilateral which effectively discriminate the facial expressions and improve the robustness to pose and illumination variations. The quadrilateral generation and trine relations are shown in Fig.2. The properties of QUEST are summarized as follows: 1) QUEST encoded the gradient edge information by dividing neighboring pixels into two quadratics to generate six-bit compact code. Thus, generates small feature vector. 2) QUEST extracted the gradient information by utilizing trine pixel relationship, that increases its robustness to noise, pose and lighting changes. 3) QUEST cohesively describe the disparities among the expression classes.

The detailed step wise representation of the QUEST is given in Eq (1-4).

$$L(x, y) = \sum_{\upsilon=0}^{p-3} \left\{ f(Q_{\upsilon,\omega} - I_c) \times 2^{\upsilon} \right\}_{w=0}^{1} \quad (1)$$

Where, $L(x, y)$ is the coded pixel for each (x, y) pixel of the I gray scale image of size $M \times N$. $I_c$ is the reference pixel in the image and $p$ is the total number of the neighborhood.

$$Q_{\upsilon,\omega} = \frac{I_{(2\upsilon-\omega+\psi(\upsilon)) \bmod p} + I_{2(\upsilon-\omega+1) \bmod p}}{(\psi(\upsilon)/3) + 2} \quad (2)$$

$$\psi(\eta) = \left(\lfloor \eta/4 \rfloor\right) \times (p-2) \quad (3)$$

$$f(x) = \begin{cases} 1 & if \ x \geq 0 \\ 0 & otherwise \end{cases} \quad (4)$$

### A. Discriminiabilty with existing methods

In literature, directional informational based descriptors like LDP, LDN, LDTexP and LDTerP have shown excellent performance in capturing salient feature of expressive regions. Innately, they extracted the directional number by applying different compass masks as sobel, kirsch and robinson. Further they represent the transitional information of the structure by mixing directional number with other parameters like texture and ternary gradient information.

Therefore, the performance of these methods fully depended on selected predesigned compass mask. However, the proposed method exploits the quad structure and extract the salient edge information of the facial appearance changes by incorporating trine pixels relationship. Thereby, it enhances

$$\chi(i, j) = \begin{cases} 1 & i = j \\ 0 & else \end{cases} \quad (6)$$

Where $\chi(i, j)$ is a resultant feature pattern, computed at (i, j) position of $\phi^{th}$ region $R_\phi$ and $p \in [0-255]$.

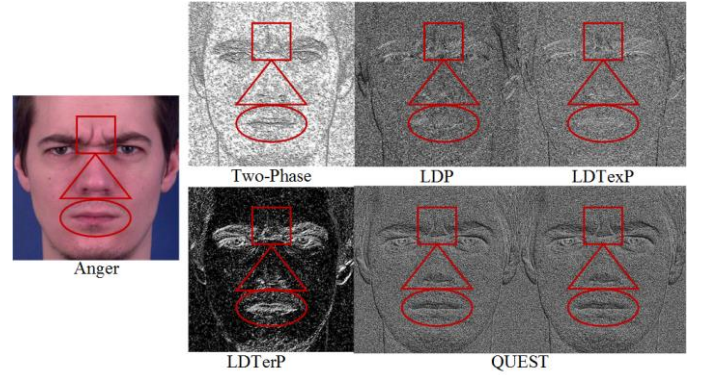

Fig. 3. The feature response images of existing and proposed descriptors over *anger* emotion class.

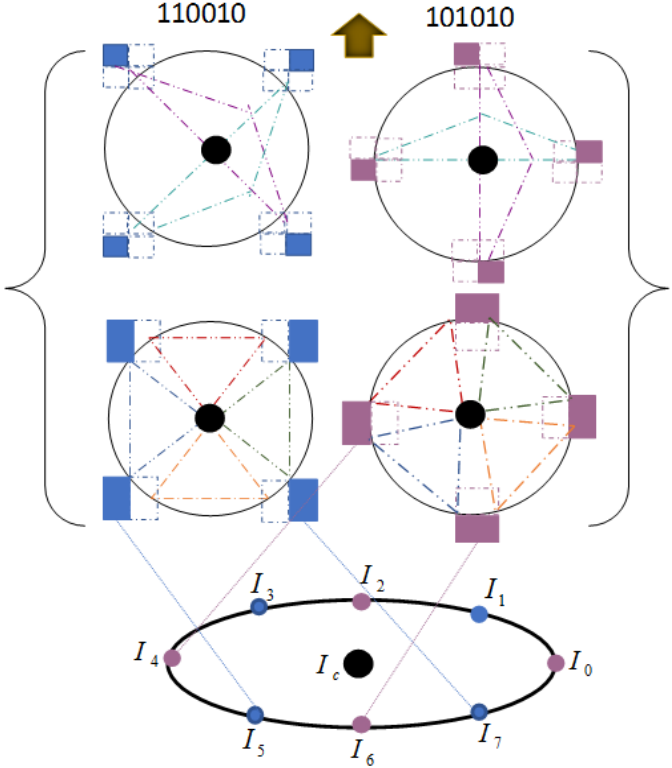

Fig. 2. The proposed QUEST descriptor

the recognition rate by identifying disparities between inter class expression. The capability of discrimination is demonstrated in Fig. 4, we have selected three different facial expression images from the fear, surprise and sad emotion class respectively. From the Fig. 4, we can see that, QUEST is able to differentiate these classes, even though these images have minor edge variations.

Thereby it has the more capability to discriminate the expression classes. Moreover, gradient information between trine pixels in particular quadratic increase the robustness against noise and illumination changes. The QUEST pattern enhances the transitional information by highlighting edge variation and avoiding the noisy elements as shown in Fig. 3.

### B. *Feature Vector & Classification*

The feature response of the QUEST is able to discriminate micro patterns of an expression region. Further, these responses are divided in N equal sized regions, $\{R_0, R_1,...R_N\}$ to preserve the minute variation of an expression region. Then, relevant feature vectors are computed by applying histogram $H_\phi$ to each region $R_\phi$. Furthermore, final feature vector is computed by adding histograms of each region through Eq. (5-6).

$$H_\phi(p) = \sum_{(i,j) \in R_\phi} \chi(f(i, j), p), \quad \forall p \quad (5)$$

In the proposed method, $N-$fold person independent validation technique was performed by utilizing Support Vector Machine (SVM) [26]. SVM is a supervised learning technique, which categorize the input data into high-dimensional data plane by applying non- linear mapping. Consequently, it generates a linear hyper-plane with maximal space to partitioned the input data into two classes in hyper-plane scope. Let $D = \{(a_i, b_i), i = 1, 2,...N\}$ be a training image set of labeled data, where $a_i \in x^n$ is the feature vector of relevant class and $b_i \in \{+1, -1\}$. A new input object is defined by Eq. (7).

$$f(t) = sign(\sum_{i=1}^{N} \gamma_i b_i k(a_i, t) + b) \quad (7)$$

Where $\gamma_i$ is the langrange's multipliers of dual optimization problem, $k$ is the SVM kernel function and $b$ implies for the bias parameter. One-against-one approach is opted to perform multi-class classification by generating $^nC_2$ binary classifiers, where n is the total number of the available class labels. Further, final result is computed by conducting voting decision technique.

The overall performance of the FER system is evaluated by accuracy rate, computed by using Eq. (8).

$$Accr. Rate = \frac{\text{Total no. of accurate predicted objects}}{\text{Total no. of objects}} \times 100 \quad (8)$$

### III. EXPERIMENTAL RESULTS AND DISCUSSION

We conduct five experiments to validate the performance of the proposed feature descriptor. Five datasets: MMI [27-28], GEMEP-FERA [29], OULU-CASIA [30] and ISED [31], are used to analyze, are used to analyze the recognition rate under different variants such as pose, ethinicity and illumination changes. Moreover, to performs the experimets , images are cropped and normalized into equal size-based on the expressive region points. In our experiments we used the Viola Jones [32] face detection algorithm, instead of manual cropping for importing a real-life application scenario to FER system.

Moreover, to generate the feature vector, response feature maps are divided into equal sized blocks $S \times S$, where $S$ =8. Thus, it provides standardized base to analyze the fair comparison of proposed work with previous methods.

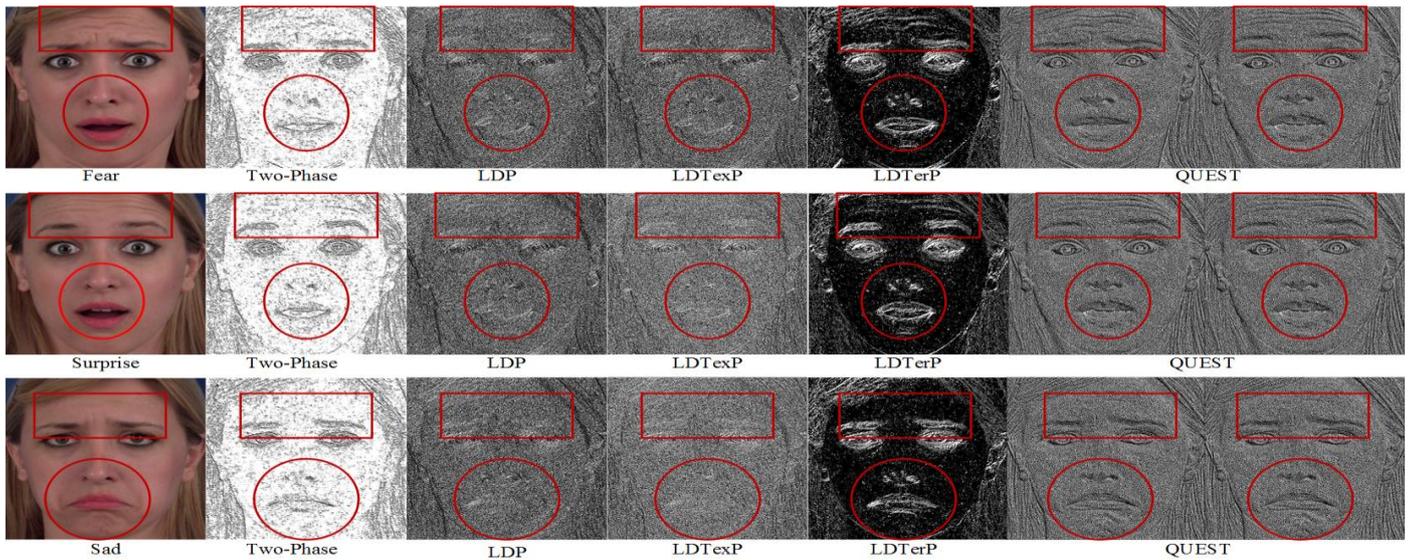

Fig. 4. The qualitative comparison between the generated response maps by applying (a) Two-Phase (b) LDP (c) LDTexP (d) LDTerP and (e) QUEST descriptors.

TABLE I
RECOGNITION ACCURACY COMPARISON ON MMI DATASET

| Methods | 6-Class Exp. | 7-Class Exp. |
| --- | --- | --- |
| LBP [11] | 76.5 | 81.7 |
| Two-Phase [13] | 75.4 | 82.0 |
| LDP [15] | 80.5 | 84.0 |
| LDN [16] | 80.5 | 83.0 |
| LDTexP [17] | 83.4 | 86.0 |
| LDTerP [18] | 80.6 | 80.0 |
| Spatio-Temopral* [25] | 81.2 | - |
| **QUEST** | **83.05** | **84.0** |

* This result is from the corresponding original paper

TABLE II
RECOGNITION ACCURACY COMPARISON ON GEMEP-FERA DATASET

| Methods | 5-Class Exp. | 6-Class Exp. |
| --- | --- | --- |
| LBP [11] | 92.2 | 87.8 |
| Two-Phase [13] | 88.6 | 85.0 |
| LDP [15] | 94.0 | 90.0 |
| LDN [16] | 93.4 | 91.0 |
| LDTexP [17] | 94.0 | 91.8 |
| **QUEST** | **94.3** | **91.33** |

For the arrangement of dataset, we have selected three to five most expressive frames from the video sequences. The former approaches widely adopted this setup [15-18]. Generally, two schemes are utilized to validate the outcomes: person dependent and person independent cross validation respectively. In person independent cross validation image set is partitioned into N folds: N-1 folds are used as training image set and rest are used as testing image set. In person dependent validation scheme, one subject's expressions are poured into test set and remaining are used for training. Our experiments are validated by using N person independent cross validation. Certain, CNN based learning techniques are also follows the same procedure to validate their results. To generates N folds, we have arranged our datasets in random 80/20 ratio and selected training and testing image sets respectively. Due to randomness in data selection, we have performed each experiment five times and average of the outcomes considered as final recognition rate.

In the following subsection, we present performance and comparative analysis of proposed method with existing approaches. We also implemented existing approaches on our experimental setup to make fair comparison between proposed and existing approaches.

### A. MMI

MMI dataset consist almost 2900 samples, which included both video and image sequences of faces posing various expressions. It includes both male and female subjects, from different region as Asia, Europe, and South America. It consists 236 sessions of 32 subjects. In MMI dataset different sessions has been taken of those subjects who wearing glasses, scarf, and cap. One session is taken with such things and another without these. Each video sequences have a start with neutral face and goes up to peak of type of its expression. We select peak expressive frames from video sequence and crop facial area of all images. Out of all the video sequences we select 561 frames in which 69 – anger, 71 – disgust, 69 – fear, 84 – happy, 67 – sad, 75 – surprise and 126 – neutral. The recognition rates are shown in Table I, over MMI dataset for both 6-class and 7-class expression as compared to other methods.

### B. GEMEP-FERA

GEMEP-FERA (Geneva multimodal emotion portrayal facial expression and recognition analysis) dataset contains 226 video sequences of 10 subjects showing different facial expressions. Multiple sessions of particular person have been recorded with 720 x 576 resolutions. In original dataset video sequences are divided into two parts: Training set consist 155 video sequences of 7 subjects and test set consist 71 video sequences of 6 subjects. Some of subjects are not recorded with all the expression class. Each image sequence labeled with one of the five expression classes: anger, fear, joy, relief, and sadness. For our experiment purpose we merge both the dataset and out of all 226 sequences we have select 212 sequences, having all five emotions. We select peak frames from each video sequence and form a dataset of 744 images consisting anger – 144, fear – 138, joy – 144, relief – 120, sad – 120 and neutral – 78. The performance of our proposed descriptor over both 5-class and 6-class datasets, shown in Table II. It is clearly visible that proposed method outperforms the existing state of the art techniques. The detailed predicted results are shown in Fig. 5 and Fig. 6.

TABLE III
RECOGNITION ACCURACY COMPARISON ON OULU-NIR DATASET FOR 6-CLASS EXPRESSIONS

| Method | Dark | Strong | Weak | Avg. |
|---|---|---|---|---|
| LBP [11] | 97.6 | 97.2 | 97.2 | 97.3 |
| Two-Phase [13] | 94.3 | 94.1 | 95.2 | 94.5 |
| LDP [15] | 96.6 | 97.5 | 97.9 | 97.3 |
| LDN [16] | 98.3 | 98.1 | 98.5 | 98.3 |
| LDTexP [17] | 98.1 | 98.0 | 98.2 | 98.1 |
| LDTerP [18] | 98.0 | 97.8 | 98.1 | 98.0 |
| **QUEST** | **98.6** | **98.2** | **98.2** | **98.3** |

TABLE IV
RECOGNITION ACCURACY COMPARISON ON OULU-NIR DATASET FOR 7-CLASS EXPRESSIONS

| Method | Dark | Strong | Weak | Avg. |
|---|---|---|---|---|
| LBP [11] | 96.4 | 96.9 | 95.9 | 96.4 |
| Two-Phase [13] | 93.0 | 92.3 | 91.3 | 92.2 |
| LDP [15] | 96.0 | 97.7 | 97.7 | 97.1 |
| LDN [16] | 96.7 | 98.1 | 98.0 | 97.6 |
| LDTexP [17] | 97.8 | 97.7 | 97.1 | 97.5 |
| LDTerP [18] | 97.7 | 96.6 | 98.2 | 98.0 |
| **QUEST** | **98.3** | **98.2** | **98.2** | **98.2** |

TABLE V
RECOGNITION ACCURACY COMPARISON ON OULU-VIS DATASET FOR 6-CLASS EXPRESSIONS

| Method | Dark | Strong | Weak | Avg. |
|---|---|---|---|---|
| LBP [11] | 94.1 | 96.3 | 96.1 | 95.5 |
| Two-Phase [13] | 80.3 | 87.8 | 90.0 | 86.0 |
| LDP [15] | 92.7 | 98.4 | 97.2 | 96.1 |
| LDN [16] | 94.3 | 98.5 | 96.0 | 96.2 |
| LDTexP [17] | 90.3 | 98.5 | 96.6 | 95.1 |
| LDTerP [18] | 93.9 | 98.3 | 97.2 | 96.4 |
| **QUEST** | **94.5** | **98.5** | **97.9** | **96.9** |

TABLE VI
RECOGNITION ACCURACY COMPARISON ON OULU-VIS DATASET FOR 7-CLASS EXPRESSIONS

| Method | Dark | Strong | Weak | Avg. |
|---|---|---|---|---|
| LBP [11] | 90.1 | 93.3 | 94.1 | 92.5 |
| Two-Phase [13] | 86.2 | 87.0 | 89.4 | 87.5 |
| LDP [15] | 94.3 | 98.0 | 96.3 | 96.2 |
| LDN [16] | 95.3 | 97.8 | 96.7 | 96.6 |
| LDTexP [17] | 95.0 | 98.3 | 96.7 | 96.7 |
| LDTerP [18] | 92.4 | 98.8 | 96.8 | 96.0 |
| **QUEST** | **94.9** | **99.1** | **97.2** | **97.0** |

### C. OULU-CASIA

OULU-CASIA dataset consist 80 subjects representing six basic expressions: anger, disgust, fear, happy, sad, surprise, aged between 23 to 58 years. All images in dataset are frontal faced. The expression images are captured with using NIR and VIS cameras. Furthermore, all images are taken in 3 different lighting conditions: strong, dark, and weak. In strong condition appropriate light is used, in weak condition only computer display light is used, while in dark condition images taken in near dark environment. Total 2880 video sequence are present in dataset, in which 480 image sequences belongs to each illumination condition. We only select 3 peak images from all the image sequence which belongs to six basic expressions. Neutral expression is created with taking initial frames of the sequences when subjects are not started showing any emotion. For our experimental purpose we arrange images in all the 3-illumination condition. The numbers of images we select in each expression as anger – 240, disgust – 240, fear – 240, happy – 240, surprise – 240 and neutral – 240. Same applied in both NIR and VIS camera images. Accuracy results computed over OULU-CASIA is tabulated in Table-III to Table VI. These results validate that, our proposed method yields better performance as compare to existing state-of-art approaches.

|  | ANG | FEA | JOY | REL | SAD |
|---|---|---|---|---|---|
| ANG | 0.91 | 0 | 0.05 | 0.03 | 0.02 |
| FEA | 0.05 | 0.93 | 0 | 0.02 | 0 |
| JOY | 0.03 | 0.01 | 0.94 | 0.02 | 0 |
| REL | 0.02 | 0.01 | 0.01 | 0.96 | 0 |
| SAD | 0.01 | 0.02 | 0 | 0 | 0.97 |

Fig. 5. Confusion matrix of QUEST for 5-class expression classification in GEMEP_FERA dataset

|  | ANG | FEA | JOY | NEU | REL | SAD |
|---|---|---|---|---|---|---|
| ANG | 0.92 | 0.04 | 0.01 | 0.02 | 0 | 0.01 |
| FEA | 0.02 | 0.87 | 0.02 | 0.04 | 0.04 | 0.01 |
| JOY | 0.01 | 0.01 | 0.96 | 0.01 | 0.01 | 0 |
| NEU | 0 | 0.04 | 0 | 0.82 | 0.03 | 0.11 |
| REL | 0 | 0.02 | 0 | 0.03 | 0.95 | 0 |
| SAD | 0.04 | 0 | 0 | 0.04 | 0.02 | 0.90 |

Fig. 6. Confusion matrix of QUEST for 6-class expression classification in GEMEP_FERA dataset

### D. ISED

ISED (Indian Spontaneous Expression Dataset) dataset contains 428 video sequences of 50 cross cultured Indian origin subjects aged from 18 to 22 years. out of all subjects 29 are male and 21 are females. The dataset consist spontaneous emotions captured without subject's awareness, in which subjects have not been forced to pose particular expression. It consists multiple video sequences of each emotion with high resolution 1920 x 1080. Each video sequence has been labeled with one of four emotions: disgust, fear, happy and fear. In our experimental setup we take onset frames of video sequence as neutral. As video sequences started with neutral and go up to peak of the particular expression, we select 3 peak frames from all the sequences. We aggregate 1186 images including all five expressions. The performance of proposed and other methods is shown in the Table VII. As results states that, propose approach outperforms existing state-of-art approaches.

## IV. CONCLUSION

In this paper, we proposed a new feature descriptor named as Quadrilateral Senary bit Pattern (QUEST) to represent the features of facial expressions. QUEST encoded six-bit compact code by thresholding neighboring pixels with reference pixel by dividing neighborhood pixels in two quadrics. QUEST extracts the transitional pattern by analyzing pixels located in quadrilaterals, that extract the edge patterns of the intensity changes in the local neighborhood. Therefore, it enhances the performance of the descriptor by capturing the salient features of expressive regions. Furthermore, Quadrilateral structure extracted only relevant features and suppress the noise, increase the robustness to noise. Experimental results show the better ability of the proposed method as it obtained better accuracy rates as compare to the state-of-art approaches over five different datasets.

TABLE VII
RECOGNITION ACCURACY COMPARISON ON ISED DATASET

| Methods | 4-Class Exp. | 5-Class Exp. |
|---|---|---|
| LBP [11] | 88.9 | 72.1 |
| Two-Phase [13] | 78.4 | 84.4 |
| LDP [15] | 88.3 | 85.6 |
| LDN [16] | 90.6 | 82.8 |
| LDTexP [17] | 91.0 | 82.8 |
| LDTerP [18] | 88.0 | 83.8 |
| **QUEST** | **91.9** | **86.2** |